\documentclass[10pt,twocolumn,letterpaper]{article}

\usepackage[pagenumbers]{iccv} 

\newcommand{\norm}[1]{\left\lVert#1\right\rVert\xspace}
\newcommand{\fnorm}[1]{\ensuremath{\|{#1}\|_\text{\scriptsize F}}\xspace} 
\newcommand{\parent}[1]{\ensuremath{\text{parent}({#1})\xspace}}
\newcommand{\cmark}{\textcolor{green!60!black}{\ding{51}}} 
\newcommand{\var}[1]{\mathit{Var}(#1)\xspace}

\renewcommand{\vec}[1]{\boldsymbol{#1}}

\definecolor{iccvblue}{rgb}{0.21,0.49,0.74}
\usepackage[pagebackref,breaklinks,colorlinks,allcolors=iccvblue]{hyperref}

\usepackage{booktabs}
\usepackage{pifont}
\usepackage{xcolor}
\usepackage{tikz}
\usepackage{bbm}
\usepackage{xspace}

\def\paperID{*****} 
\def\confName{ICCV}
\def\confYear{2025}

\title{Geometry-Aware Losses for Structure-Preserving Text-to-Sign Language Generation}

\author{Zetian Wu\\
{\tt\small wuzet@oregonstate.edu}
\and
Tianshuo Zhou\\
{\tt\small zhoutian@oregonstate.edu}
\and
Stefan Lee\\
{\tt\small leestef@oregonstate.edu}
\and
Liang Huang\\
{\tt\small liang.huang@oregonstate.edu}\\
Oregon State University\\
}

\begin{document}
\maketitle
\begin{abstract}

Sign language translation from text to video plays a crucial role in enabling effective communication for Deaf and hard-of-hearing individuals. A major challenge lies in generating accurate and natural body poses and movements that faithfully convey intended meanings. Prior methods often neglect the anatomical constraints and coordination patterns of human skeletal motion, resulting in rigid or biomechanically implausible outputs. To address this, we propose a novel approach that explicitly models the relationships among skeletal joints—including shoulders, arms, and hands—by incorporating geometric constraints on joint positions, bone lengths, and movement dynamics. During training, we introduce a parent-relative reweighting mechanism to enhance finger flexibility and reduce motion stiffness. Additionally, bone-pose losses and bone-length constraints enforce anatomically consistent structures. Our method narrows the performance gap between the previous best and the ground-truth oracle by 56.51\%, and further reduces discrepancies in bone length and movement variance by 18.76\% and 5.48\%, respectively, demonstrating significant gains in anatomical realism and motion naturalness.
\end{abstract}    
\section{Introduction}
\label{sec:intro}
Sign language translation from text to video plays a crucial role in enabling effective communication for Deaf and hard-of-hearing individuals by serving as a vital bridge to the text-dominated digital world \cite{zhou+:2022spatial, armstrong+:2012origins, boyes+:2001hands, campbell+:2007sign, ong+:2005automatic, hickok+:1996neurobiology}. For many native signers, processing text requires a significant cognitive load, increasing the likelihood of misinterpretation. By converting written information into a visual-gestural format, these translation systems aim to provide equitable access to digital content and services, ensuring that meaning is conveyed in a user's native linguistic framework \cite{robert+:2011sign, sandler:2012phonological}.

A major challenge in this domain lies in generating accurate and natural body poses and hand gestures that faithfully convey intended meanings. While broader body movements like torso shifts and head tilts provide crucial grammatical context, the primary difficulty lies in the precise and fluid articulation of the hands. The human hand is an articulator with exceptionally high degrees of freedom, and sign language exploits this complexity to encode a vast lexicon. Subtle variations in handshape, palm orientation, and the trajectory of movement can fundamentally alter a sign's meaning, making the synthesis of these fine-grained, rapid motions a significant technical hurdle. Therefore, while full-body kinematics are necessary for naturalness, mastering the intricate dynamics of the hands remains the paramount challenge for achieving linguistic accuracy.

Among early influential approaches, \citet{saunders+:2020progressive} introduced the Progressive Transformer (Sec.~\ref{sec:prelim}), a widely cited framework in the field that laid foundational techniques for continuous sign language generation. Although more recent methods have since surpassed its performance, it remains an important baseline for evaluating improvements. In this work, we build upon the Progressive Transformer’s basic structure by explicitly modeling the relationships among skeletal joints, including the shoulders, arms, and hands, by incorporating geometrical constraints on their positions, lengths, and movements. Specifically, we propose parent-relative reweighting in Sec.~\ref{sec:adp}, a mechanism designed to enhance the flexibility of finger movements, which addresses the stiffness commonly observed in previous methods. Furthermore, we integrate bone-pose losses and bone-length constraints in Sec.~\ref{sec:geo} to enforce anatomically consistent motion while maintaining natural positioning. Finally, while the Progressive Transformer employs counter embeddings for controlling frame termination—which often results in unstable predictions—we propose a stable, explicit end-of-sequence (EOS) control mechanism detailed in Sec.~\ref{sec:eos}, further enhancing the coherence and quality of generated sign sequences. 

To evaluate our approach, we conduct experiments (Sec.~\ref{sec:exp}) on the PHOENIX-2014 dataset \cite{koller+:2015continuous}, a widely used benchmark for sign language translation. Our method achieves a BLEU-4 score of 14.22, reducing the performance gap between the previous best method and the theoretical upper bound (back-translation on ground truth) by 56.51\%. This improvement highlights the effectiveness of our geometrical constraints in producing more natural and accurate sign language videos. Additionally, we reduce the gap between predictions and references in bone length and movement variance by 18.76\% and 5.48\%, respectively, further demonstrating the anatomical consistency achieved by our model.

Overall, our contributions in this work are threefold: (1) We propose an parent-relative reweighting mechanism (Sec.~\ref{sec:adp}) to enhance the flexibility and naturalness of finger movements, addressing the stiffness typically observed in existing approaches; (2) We introduce explicit geometric constraints—including bone-pose losses and bone-length constraints(Sec.~\ref{sec:geo})—to ensure anatomically consistent skeletal motion throughout the generated sequences; and (3) We present a stable end-of-sequence (EOS) control mechanism (Sec.~\ref{sec:eos}) that significantly improves the stability and coherence of sequence termination compared to prior methods. Our comprehensive experimental evaluation on the PHOENIX-2014 benchmark dataset demonstrates substantial improvements in translation quality, motion realism, and anatomical consistency, advancing the state of the art in text-to-sign video generation.
\section{Preliminaries} 
\label{sec:prelim}
\begin{figure}
  \centering
   \includegraphics[width=\linewidth]{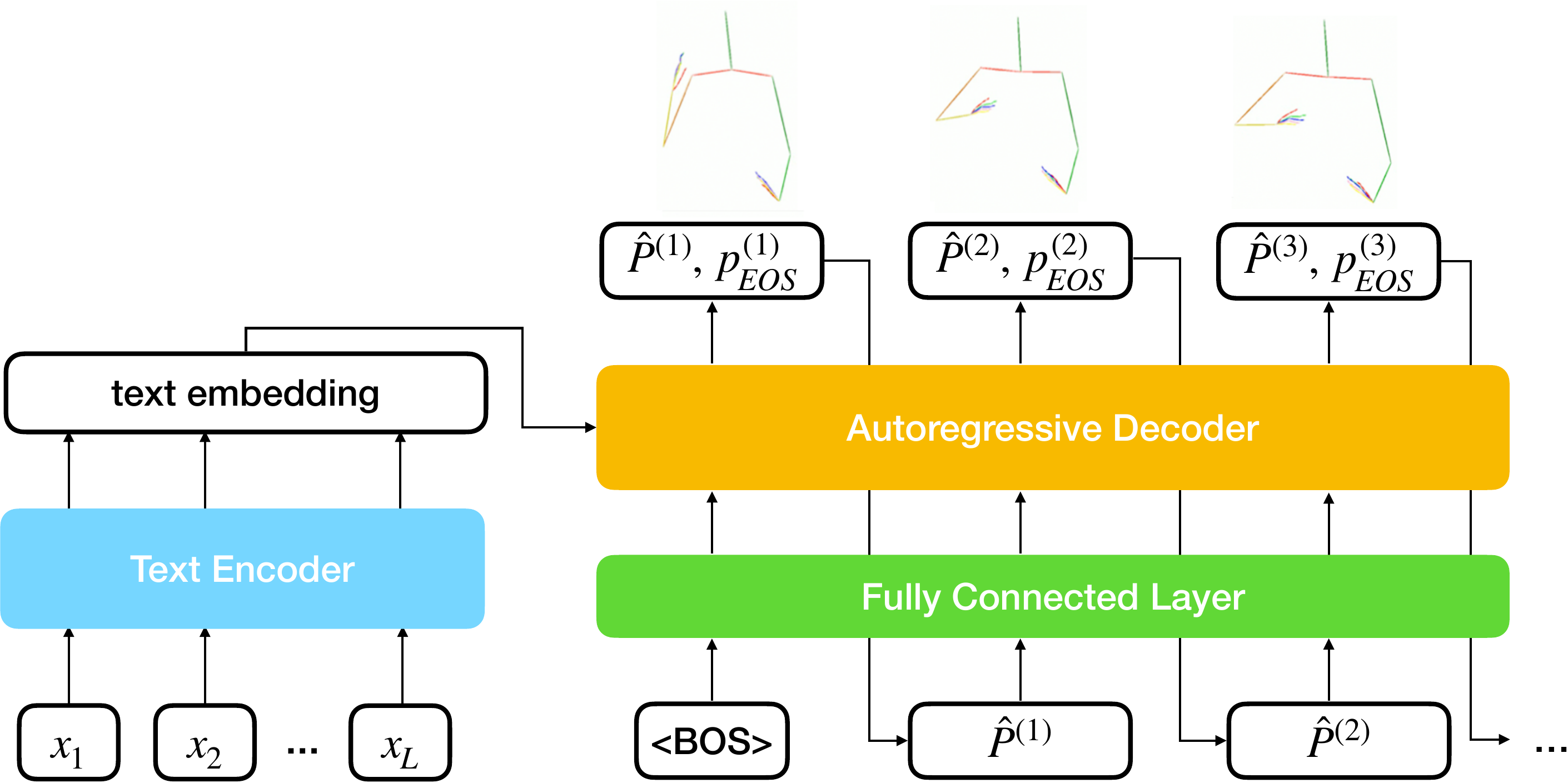}

   \caption{Model overview. This model inputs text sequences $\boldsymbol{x}$ and outputs skeleton sequences $\boldsymbol{P}$ autoregressively. We also use the hidden embedding from the last layer of the decoder to predict the end of generation.}
   \label{fig:encoder_decoder}
\end{figure}

To enhance readability, we first introduce the Progressive Transformer (PT)~\cite{saunders+:2020progressive}, the prior approach upon which our work builds. We provide an overview of its model, loss function, and counter-decoding technique. We also present our final model architecture in Fig.~\ref{fig:encoder_decoder}.

PT is the first end-to-end Sign Language Production (SLP) method that translates spoken language text into sign pose sequences. Each pose frame indexed by $t$ consists of a set of N joints, represented as a matrix $P^{(t)}=[\boldsymbol{p}_0, \boldsymbol{p}_1, \dots, \boldsymbol{p}_{N-1} ]$, where each joint $\boldsymbol{p}_i = [x, y, z]^T (i=0, 1, \dots, N-1$) is a 3D coordinate. The PT model follows a standard encoder-decoder transformer architecture. The encoder maps input symbols (text or gloss) into a latent representation, while the decoder autoregressively generates a sequence of sign language skeleton poses, attending to the encoded features. Additionally, a decoding counter, valued between 0 and 1, is learned alongside the skeleton coordinates to capture temporal progression and control the end of sequence generation. In essence, PT can be regarded as an autoregressive predictor, i.e.,
\begin{equation}
    \hat{P}^{(t)}, \hat{c}^{(t)} = \ensuremath{\text{PT}}_{\theta}(\boldsymbol{x}, \hat{P}^{(<t)}, \hat{c}^{(<t)}),
\end{equation}
where $\boldsymbol{x}$ denotes the input spoken language tokens, $t$ is the index of pose frames, $c$ represents the decoding counter, and $\theta$ denotes the parameters of PT. As pointed out by the previous work NAT-EA~\cite{huang+:2021towards}, the original implementation of PT used ground-truth information including the first pose frame and the decoding counter ($c$) at inference, which is not realistic. In our reproduction of PT, we follow the setting of NAT-EA~\cite{huang+:2021towards} by removing the first ground-truth pose frame and using the predicted decoding counter ($\hat{c}$) at inference.

The loss function in the training process is based on the Mean Squared Error (MSE) between predicted and ground truth coordinates, 
\begin{equation}
    \mathcal{L}_{\text{PT}}(\theta) = \sum_{t} \fnorm{{P^{(t)}-\hat{P}^{(t)}}},
\end{equation}
where \fnorm{\cdot} is the Frobenius norm of a matrix.
Data augmentation techniques, such as adding Gaussian noise during training and a multi-frame generation strategy, further enhance PT’s performance.

Despite its success in demonstrating end-to-end training for SLP, PT has several limitations. First, its learning objective disregards the geometric structure and constraints of the human body, treating each pose as an independent set of coordinates. Second, it relies on ground truth decoding counters during inference rather than the predicted counters, raising concerns about their actual contribution to autoregressive generation. Finally, while back translation is commonly used to evaluate SLP, it may be insufficient, as even ground truth poses yield low BLEU scores. Therefore, additional intermediate metrics are needed to better assess sign pose generation and improve SLP system evaluation.

\section{Parent-relative Reweighting}
\label{sec:adp}
Parent-relative reweighting is designed to enhance the flexibility of finger movements in sign language generation by dynamically adjusting the influence of different skeletal joints during training. Standard approaches \cite{saunders+:2020progressive, saunders+:2020adversarial, saunders+:2021mixed, huang+:2021towards, hwang+:2021non, hwang+:2024gloss, zhou+:2019dance} often treat all joints equally, leading to an imbalance where larger joints (e.g., shoulders, elbows) dominate optimization while fine-grained finger articulation remains underrepresented. To address this, parent-relative weighting introduces a joint-specific weighting mechanism that prioritizes critical articulators. \looseness=-1

Specifically, we define each link as the relative position between a joint and its parent joint: $\vec{b}_i = \vec{p}_i - \vec{p}_{\parent{i}}$ to denote each link. The lower neck is taken as the root joint. We then compute the spatio-temporal variance for each joint across all frames, denoted as $\sigma_i^2$. For the root joint, we calculate variance based on its absolute position, $\sigma^2_{root} = \var{\vec{p}_{root}}$, whereas for all other joints, we compute variance based on their relative positional differences: $\sigma^2_i = \var{\vec{b}_i}$ for $i$ is not the root. The weighting for each joint is then determined as:
\begin{equation}
    w_i \;=\; 1 \;-\; \frac{\sigma_i^2}{\sum_i\sigma_i^2}
\end{equation}
Our approach is inspired by \citet{hwang+:2025spatio}, which computed variance based on the absolute joint position sequence or partial relative joint position, e.g. hand joints v.s. wrist.  While this approach captures overall movement trends, it does not account for the inherent structural dependencies between joints. In contrast, our work inherently normalizes for global translation and better captures the local motion dynamics within the skeletal hierarchy by leveraging the relative representation. This distinction allows our method to emphasize joint-wise articulation patterns rather than being influenced by absolute spatial variations.

This weighting is integrated into the pose reconstruction loss, where each joint’s contribution is scaled by its assigned weight, encouraging the model to focus on expressive and flexible hand movements while maintaining overall anatomical consistency. By adaptively modulating joint influence, APW mitigates the rigidity observed in prior methods, leading to more natural and fluid sign language motion. \looseness=-1
\section{Geometrical losses}
\label{sec:geo}
Following \citet{saunders+:2020progressive} we use Mean Squared Error (MSE) Loss to minimize the discrepancy between predicted and ground truth joint positions:
\begin{equation}
    L_{\text{MSE}} = \fnorm{P-\hat{P}},
\end{equation}
where \fnorm{\cdot} denotes the Frobenius norm. While MSE loss encourages accurate joint localization, it does not explicitly constrain the skeletal structure. As a result, we observed that the generated skeletons tend to be shorter than the ground truth due to the lack of direct enforcement on bone lengths.

To address this issue, we follow and incorporate Bone Length Loss, which explicitly constrains bone magnitudes to maintain realistic skeletal proportions. However, while this loss ensures length consistency, it does not account for the relative motion of bones, which is crucial for capturing dynamic skeletal movements. To address this, we propose Bone Pose Loss, which explicitly enforces consistency in how bones move over time. Below, we detail these two additional loss functions.

\subsection{Bone Length Loss}
Although MSE loss encourages accurate joint placement, it does not directly regulate the lengths of bones, leading to potential skeletal shrinkage. To mitigate this issue, we incorporate Bone Length Loss, following \citet{matsune+:2024geometry}, to enforce bone length consistency. This loss is defined as:
\begin{equation}
    L_{\text{bone}} = \sum_i\lambda_i\frac{\left|{\norm{\vec{b}_i}_2-\|{\hat{\vec{b}_i}\|_2}}\right|}{\|\vec{b}_i\|_2}
\end{equation}
where $\vec{b}_i$ represents the relative position of a joint with respect to its parent, i.e. link, as introduced above. By penalizing deviations in bone lengths while normalizing with respect to the ground truth, this loss ensures that the predicted skeleton maintains natural proportions. The weighting factor $\lambda_i$ is determined based on the bone-length difference observed in the development set. Specifically, we calculate $\lambda_i$ as:
\begin{equation}
    \lambda_i = \frac{1}{|\mathcal{D}|} \sum_{j} \frac{\left|\|\vec{b}_i^j\|_2 - \|\hat{\vec{b}}_i^j\|_2\right|}{\|\vec{b}_i^j\|_2}
\end{equation}
where $\mathcal{D}$ denotes the development set, and $\vec{b}_i^j$, $\hat{\vec{b}}_i^j$ represents the predicted and true length of $i$-th bone of $j$-th sample in the development set, respectively. This formulation ensures that bones with larger discrepancies in the development set receive higher weighting, enforcing stronger length constraints where necessary.

With this loss formulation, we effectively prevent skeleton shrinkage and ensure structural stability in the predicted skeletons. \looseness=-1

\subsection{Bone Pose Loss}
While Bone Length Loss preserves skeletal proportions, it does not ensure that bones exhibit consistent motion patterns. To address this, we propose Bone Pose Loss, which enforces smooth and natural temporal dynamics in bone movement. This loss is formulated as:
\begin{equation}
    L_{\text{pose}} = \sum_{i} \frac{\|\vec{b}_i-\hat{\vec{b}}_i\|_{2}}{\|\vec{b}_i\|_2 }
\end{equation}
This loss penalizes discrepancies in bone movement trajectories, ensuring that the predicted skeleton not only maintains accurate bone lengths but also exhibits realistic movement patterns. By enforcing smooth and consistent bone motion, this loss helps capture natural skeletal dynamics.
\section{End-of-sequence Control}
\label{sec:eos}
\citet{saunders+:2020progressive} employed Counter Embedding (CE) to regulate the end of generation, where a counter value between 0 and 1 represents the frame’s relative position within the total sequence length. While this approach provides a temporal reference, the prediction itself remains unstable, which in turn causes inconsistencies in determining the end of the sequence.

To address this issue, we replace the counter embedding with a more explicit End-of-Sequence (EOS) Embedding. Instead of relying on a continuous positional value, we introduce a binary embedding that directly indicates whether the generation should continue or stop. 

To implement EOS control, we introduce a learned decision mechanism that predicts whether the sequence should end at each time step. Specifically, we take the final hidden representation of the current token, denoted as $\vec{h}^{(t)}$, and pass it through a linear layer followed by a sigmoid activation to obtain an end-of-sequence probability: \looseness=-1
\begin{equation}
    p_{\text{EOS}} = \sigma(\vec{W}_{\text{EOS}} \vec{h}^{(t)} + \vec{B}_{\text{EOS}})
\end{equation}
where:
\begin{itemize}
    \item $\vec{W}_{\text{EOS}}$ and $\vec{B}_{\text{EOS}}$ are learnable parameters of the EOS classification layer
    \item $\vec{h}^{(t)}$ is the last layer hidden embedding of current frame $t$
    \item $\sigma(\cdot)$ is the sigmoid activation function, ensuring the output remains in the range $(0,1)$
\end{itemize}
We then apply a threshold to determine whether the sequence should terminate:
\begin{equation}
    \hat{e}_{\text{EOS}} = \mathbbm{1} [p_{\text{EOS}} > \tau]
\end{equation}
where $\mathbbm{1}(\cdot)$ is the indicator function, and $\tau$ is a predefined threshold (e.g., 0.5). If $\hat{e}_{\text{EOS}} = 0$, the generation process stops; otherwise, it continues.

With the introduction of EOS embedding, the model explicitly learns when to stop generation based on the overall sequence context. This approach eliminates the reliance on implicit counter-based cues, reducing prediction instability and ensuring more reliable and consistent sequence termination. By leveraging the BOS token’s final hidden state, the model incorporates global sequence-level information to make an informed decision about when to end, leading to smoother and more natural motion synthesis.
\section{Experiments}
\label{sec:exp}
In this section, we describe our experimental setup and present the evaluation results. We conduct experiments using the PHOENIX14T dataset, a widely used benchmark for continuous sign language translation (SLT) introduced by \citet{koller+:2015continuous}. Derived from German weather forecast recordings, PHOENIX14T provides parallel sign language videos and German text translations with refined segmentation boundaries. The dataset consists of 8,257 videos performed by 9 different signers, covering a vocabulary of 2,887 German words and 1,066 unique glosses, totaling 835,356 frames. Its structured annotations and alignment make it ideal for evaluating sign language translation models. \looseness=-1

To process the data, we follow the baseline pipeline to convert sign videos into 3D skeleton pose sequences. We first extract 2D joint positions using OpenPose \cite{zhe+:2019openpose,simon+:2017hand,wei+:2016convolutional,cao+:2017realtime}, which are then lifted to 3D using a skeletal model estimation approach \cite{zelinka+:2020neural}. An iterative inverse kinematics method refines the 3D pose to maintain bone length consistency and correct misaligned joints. Finally, skeleton normalization \cite{stoll+:2018sign} ensures uniform scaling, and each joint is represented in $(x, y, z)$ coordinates.

In our experiments, we modify the encoder in \citet{saunders+:2020progressive} by replacing with the pretrained XLM-R model \cite{conneau+:2020unsupervised}. The original encoder lacks linguistic richness, limiting its ability to effectively capture the relationship between sign motion and spoken language. By incorporating XLM-R, a multilingual transformer-based encoder trained on a diverse range of textual data, we enhance the model’s ability to generate more semantically accurate sign sequences. This modification allows the system to better align motion representations with textual meanings, ultimately improving translation performance.

For evaluation, we assess the generated sequences across several key aspects. Back translation BLEU is used to measure semantic accuracy by translating generated sign sequences back into spoken language. Bone length consistency evaluates whether the model maintains stable skeletal proportions, preventing unnatural deformations. Movement accuracy assesses how well the generated motion aligns with natural movement patterns. Finally, frame length stability examines the reliability of sequence duration to ensure a coherent temporal structure.

By incorporating these evaluation metrics, we provide a thorough assessment of the model’s performance to generate accurate, natural, and temporally stable sign language sequences. \looseness=-1

\subsection{Qualitative Evaluation}
\begin{figure}
  \centering
   \includegraphics[width=\linewidth]{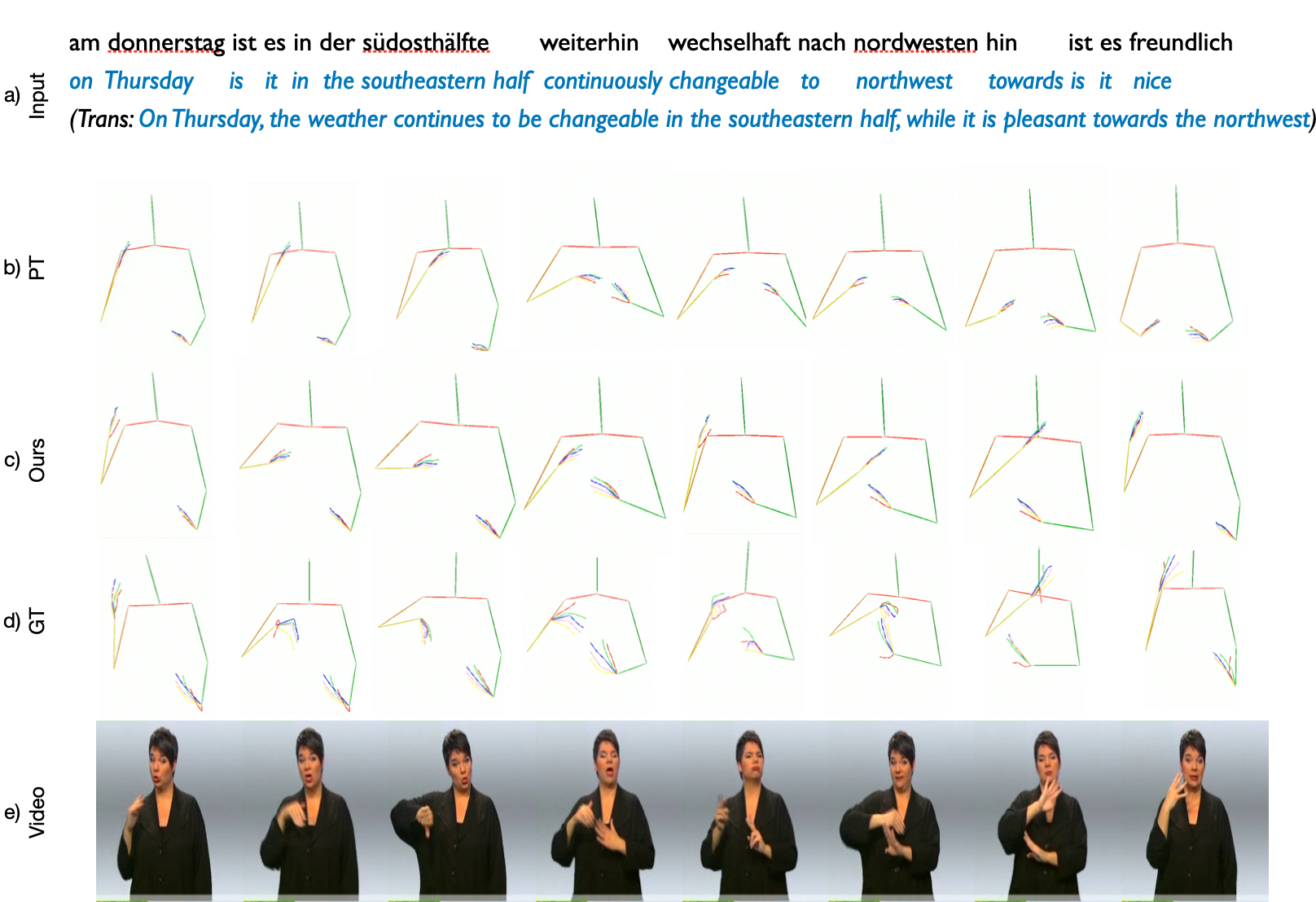}

   \caption{Qualitative evaluation of an example sign pose sequence. The source input is at the top, with the ground truth video frames and poses at the bottom. Middle rows contain produced sign pose sequences of PT and our method.}
   \label{fig:qualitative}
\end{figure}

To qualitatively evaluate our approach, we visually compare representative keyframes generated by our model with those produced by the progressive transformer baseline in Fig.~\ref{fig:qualitative}. Our method consistently generates anatomically accurate and natural hand shapes, finger articulations, and arm positions, effectively capturing the nuanced movements crucial for expressive sign language. In contrast, the progressive transformer often produces ambiguous hand shapes, unnatural finger positions, and inconsistent arm movements. Additionally, our generated poses exhibit greater consistency in bone lengths compared to the baseline, closely aligning with the anatomical proportions observed in ground-truth data. These visual comparisons highlight our method’s enhanced realism, temporal coherence, and anatomical correctness, significantly improving the clarity and interpretability of sign language sequences.

\subsection{Back translation results}
To evaluate the effectiveness of our proposed methods, we conduct experiments using sign language transformer \cite{camgoz+:2020sign} as the back translation model following \citet{saunders+:2020progressive} for consistency and calculating BLEU scores. We first perform the ablation study to analyze the contribution of each component, followed by a comparison between our final model and the baseline.

In Table~\ref{tab:ablation}, we list both reproduction results of \citet{saunders+:2020progressive} and those using XLM-R as the baseline. The ablation study systematically introduces different components to the baseline model and examines their impact on translation performance. The baseline model provides a foundation, but we observe consistent improvements as additional methods are incorporated. Parent-relative reweighting enhances joint dependencies, while bone-length loss ensures structural consistency. Bone-length reweighting and bone-pose loss further refine motion accuracy, and replacing the counter embedding with a sigmoid-based termination mechanism stabilizes sequence generation. 

After verifying the effectiveness of each component, we evaluate the final model that integrates all proposed methods. The last row in Table~\ref{tab:ablation} presents the results, showing that our approach narrows the discrepancy with the ground truth by 56.51\% compared to the baseline. The enhancements in structural constraints, bone-length regulation, and sequence termination contribute to better translation quality, demonstrating the advantage of incorporating motion-aware adjustments in sign generation.

\begin{table*}[t]
    \centering
    \renewcommand{\arraystretch}{1.2}
    \setlength{\tabcolsep}{1pt}
    \resizebox{\textwidth}{!}{
    \begin{tabular}{l ccccc cccccc}
        \toprule
        \multicolumn{1}{c}{\textbf{Configuration}} & 
        \multicolumn{5}{c}{\textbf{Methods}} & 
        \multicolumn{5}{c}{\textbf{Results on Dev}} \\ 
        \cmidrule(lr){2-7} \cmidrule(lr){8-12}
        
        & XLM-R & Parent-Rel. & Bone-Len. & Bone-Len. & Bone-Pose & Sigmoid & BLEU-1 & BLEU-2 & BLEU-3 & BLEU-4 \\
        &  & Reweighting & Loss & Reweighting & Loss &  & $\uparrow$ & $\uparrow$ & $\uparrow$ & $\uparrow$ \\
        \midrule
        Ground-Truth & - & - & - & - & - & - & \textbf{36.22} & \textbf{24.78} & \textbf{19.52} & \textbf{16.59} \\
        Reproduction & - & - & - & - & - & - & 26.48 & 15.36 & 11.50 & 9.52 \\
        \midrule
        Baseline & \cmark & - & - & - & - & - & 27.88 & 17.01 & 12.92 & 11.14 \\
        & \cmark & \cmark & - & - & - & - & 28.23 & 17.62 & 12.75 & 12.07 & (+0.93) \\
        & \cmark & - & \cmark & - & - & - & 28.38 & 17.84 & 14.09 & 12.53 & (+1.39) \\
        & \cmark & - & \cmark & \cmark & - & - & 28.47 & 17.95 & 14.21 & 12.72 & (+1.58) \\
        & \cmark & - & - & - & \cmark & - & 28.22 & 17.58 & 12.69 & 12.01 & (+0.87) \\
        & \cmark & - & - & - & - & \cmark & 28.01 & 17.17 & 13.02 & 11.56 & (+0.42) \\
        & \cmark & \cmark & \cmark & \cmark & \cmark & \cmark & 30.91 & 20.07 & 16.02 & 14.22 & (+3.08) \\
        \bottomrule
    \end{tabular}
    }
    \caption{Ablation study evaluating the impact of different methods on BLEU scores. $\uparrow$ indicates higher values are better.}
    \label{tab:ablation}
\end{table*}

\subsection{Bone length comparison}

\begin{figure}
  \centering
   \includegraphics[width=\linewidth]{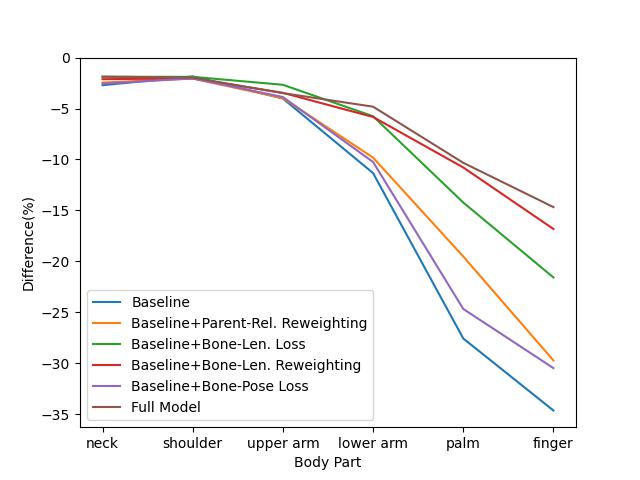}

   \caption{Average bone length deviation (\%) across different body parts.}
   \label{fig:bone_length}
\end{figure}

We assess the effectiveness of bone-length constraints by measuring the average deviation of generated bone lengths from the ground truth. As shown in Figure~\ref{fig:bone_length}, bones are categorized into six groups from top to bottom: neck, shoulder, upper arm, lower arm, palm, and finger.

The baseline results reveal a trend where bones farther from the root (neck) exhibit greater discrepancies from the ground truth. With our proposed methods, this deviation is significantly reduced, leading to a more uniform distribution of bone-length accuracy and an overall reduction in such differences. \looseness-1

\subsection{Movement comparison}
To quantitatively assess the quality of generated movements, we computed the velocity and variance of movements both globally and locally.

The variance measures the spatial variability of joint positions across all frames within a sequence. A realistic model should exhibit variance comparable to the ground truth, reflecting natural, expressive motion dynamics. The global position variance for a joint $i$ is defined as $\sigma^2_{\text{global}}(i) = \var{\vec{p}_i}$ while the local position variance $\sigma^2_{\text{local}}(i) = \var{\vec{p}_i-\vec{p}_{\parent{i}}}$, computed relative to the parent joint $\parent{i}$, is the same as $\sigma^2_i = \var{\vec{b}_i}$ in Sec.~\ref{sec:adp}. 

Similarly, the velocity, denoted as $\nu$ evaluates the smoothness and continuity of motion, capturing how naturally each joint moves from frame to frame. A realistic sequence should demonstrate movement patterns consistent with natural sign language transitions. Globally, this metric is defined as:
\begin{equation}
\nu_{\text{global}}(i) = \frac{1}{T-1}\sum_{t=1}^{T-1}\|\vec{p}_i^{(t+1)} - \vec{p}_i^{(t)}\|_2
\end{equation}

Correspondingly, the local average movement per frame, considering joint movements relative to their parent joint, is computed as:
\begin{equation}
\nu_{\text{local}}(i) = \frac{1}{T-1}\sum_{t=1}^{T-1}\|\vec{b}_i^{(t+1)}- \vec{b}_i^{(t)}\|_2
\end{equation}

\begin{figure}
  \centering
   \includegraphics[width=\linewidth]{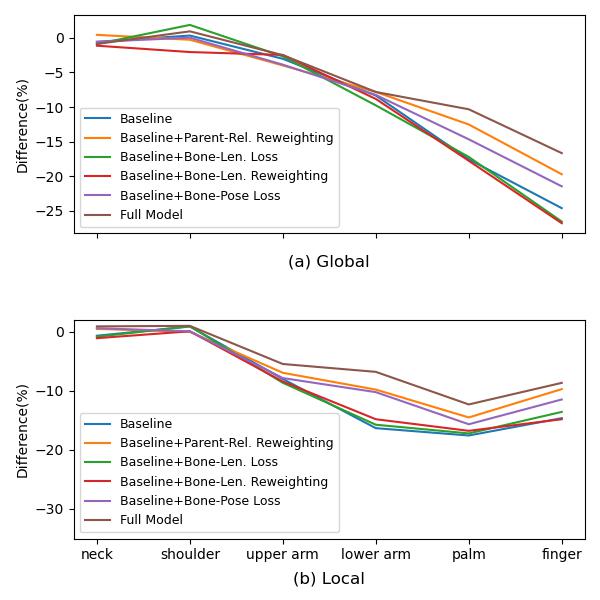}

   \caption{Average movement variance deviation (\%) across different body parts. (a) shows the absolute position variance. (b) shows the relative position variance.}
   \label{fig:move_var}
\end{figure}

\begin{figure}
  \centering
   \includegraphics[width=\linewidth]{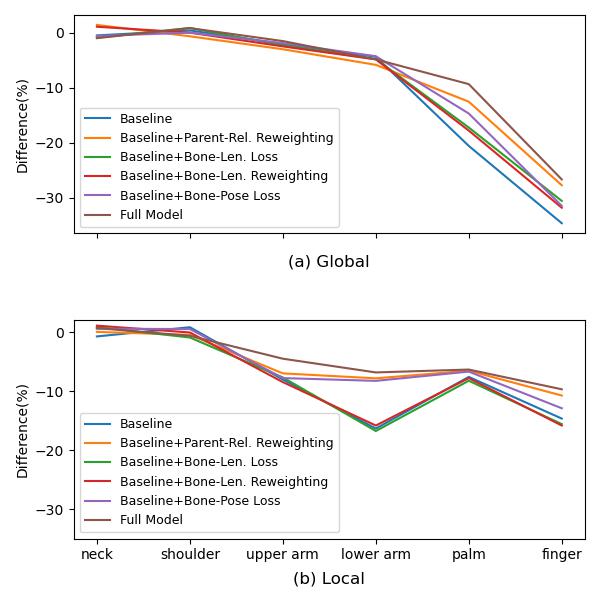}

   \caption{Average movement velocity deviation (\%) across different body parts. (a) shows the absolute position velocity. (b) shows the relative position velocity.}
   \label{fig:move_vel}
\end{figure}

Fig.~\ref{fig:move_var} and Fig.~\ref{fig:move_vel} illustrate the position variance and velocity differences between predictions and ground truth, respectively. By analyzing these metrics both globally and locally, we comprehensively assess our model’s capability to generate realistic, smooth, and anatomically consistent movements. The results highlight that parent-relative re-weighting substantially improves model performance, contributing significantly to more natural joint articulations and motion smoothness. In contrast, the introduction of bone-pose constraints yields a modest but consistent improvement. Together, these strategies demonstrate clear efficiency in capturing nuanced movements, ensuring anatomically plausible and temporally coherent sign language generation. \looseness=-1

\subsection{Frame length comparison}
\begin{figure}
  \centering
   \includegraphics[width=\linewidth]{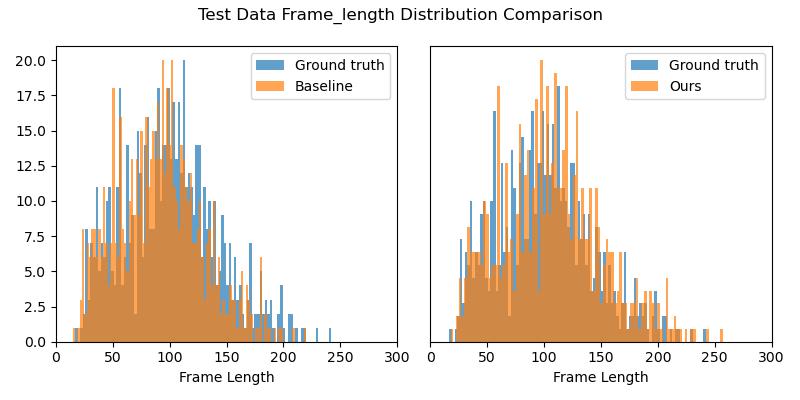}

   \caption{Frame length distribution histogram comparison on test data. Average frame length for ground truth, baseline and full model prediction is 100, 91 and 106 respectively.}
   \label{fig:data_dis}
\end{figure}

We evaluated the accuracy of generated sequence lengths by examining the relative differences in frame lengths between predictions and ground truth. Compared to the baseline with counter embedding, which underestimated lengths by 8.84\%, our proposed method yields a slight overestimation of 6.32\%. The absolute improvement is thus 2.52\% More importantly, as illustrated in Fig.~\ref{fig:data_dis}, our method produces a frame-length distribution substantially more similar to the ground truth, highlighting improved accuracy in predicting sequence lengths.

\section{Related Work}
\label{sec:related}
\subsection{Sign Language Production}
Sign Language Translation (SLT) is to translate sign languages into spoken languages, e.g., SLT from German Sign Language (DGS) to German or from American Sign Language (ASL) to English. Sign Language Production (SLP) is the inverse problem of SLT, i.e., SLP aims to convert spoken languages to sign languages with the same meaning. The output signs of SLP can be represented by skeleton poses~\cite{saunders+:2020progressive}, realistic photos~\cite{saunders+:2022signing}, 3D avatars~\cite{baltatzis+:2024neural}, etc.

Progressive Transformer (PT)~\cite{saunders+:2020progressive} claimed to be the first end-to-end Sign Language Production (SLP) method that translates spoken language text into sign pose sequences (See Section~\ref{sec:prelim} for details). One following-up work~\cite{saunders+:2020adversarial}
enhanced PT by incorporating a discriminator into the generator, improving the realism of sign production. Another subsequent work, MoMP~\cite{saunders+:2021mixed}, further enhanced the expressiveness of PT-based sign animations by learning motion primitives at the frame-level using Mixture of Experts (MoEs). In addition, NAT-EA~\cite{huang+:2021towards} and NSLP-G~\cite{hwang+:2021non} have designed non-autoregressive decoders for sign sequence generation to address the issue of error propagation in autoregressive models. Recently, UniGloR~\cite{hwang+:2025spatio} was proposed to learn a latent representation by self-supervised learning in order to capture the spatio-temporal features of sign languages, and such representation can be integrated into both SLT and SLP. \looseness=-1

Beyond end-to-end SLP methods, various approaches focus on two-stage or gloss-dependent SLP~\cite{stoll+:2018sign}. SLP is practically decomposed into two subprocesses~\cite{stoll+:2018sign}: (1) Text2Gloss translates spoken language sentences into sign gloss sequences, and (2) Gloss2Pose or Gloss2Sign learns a mapping from glosses to sign poses. For instance, Frame Selection Networks (FS-NET)~\cite{saunders+:2022signing} were designed to learn the temporal alignment between intermediate dictionary (sign) sequences and the final continuous sign sequences, with a separate SignGAN~\cite{saunders+:2022signing} model converting sign skeletons into realistic images. Other notable two-stage or gloss-dependent SLP methods include G2P-DDM\cite{xie+:2024g2p} and SignVQNet~\cite{hwang+:2024gloss}. However, we omit their details as our work focuses on advancing gloss-free end-to-end SLP.

\subsection{Text to Motion Generation}
SLP shares similarities with text-to-motion generation tasks in terms of input-output modalities. Early text-to-motion generation studies applied sequence-to-sequence models using recurrent neural networks (RNNs) for mapping textual descriptions to corresponding motions~\cite{lin+:2018generating, plappert+:2018learning}. A two-stream hierarchical encoder-decoder architecture~\cite{ghosh+:2021synthesis} was proposed to explore a finer joint-level mapping between natural language sentences and 3D pose sequences corresponding to the given motion. TM2T~\cite{guo+:2022tm2t} introduced a novel motion representation-motion token, which facilitates downstream neural machine translators (NMTs) to construct mappings between texts and motions. MotionCLIP~\cite{tevet+:2022motionclip} is a transformer-based motion auto-encoder that is trained to reconstruct motion while being aligned to the latent space of CLIP~\cite{radford+:2021learning} to leverage the generalizable representation of CLIP. 

More recently, various generative models have also been exploited to advance text-to-motion research. To enhance the diversity of generated motions, a two-stage approach~\cite{guo+:2022generating} was proposed which contains a text2length sampling and a text2motion generation module. Text2motion engages the variational autoencoder (VAE) to generate motions of various lengths sampled from the text2length stage. MDM~\cite{tevet+:2022human} introduced a diffusion-based human motion generative model with applications of text-to-motion, action-to-motion, and unconditioned generation. MLD~\cite{chen+:2023executing} demonstrated the effectiveness of diffusion models on latent spaces instead of raw motion sequences. T2M-GPT~\cite{zhang+:2023generating} investigated a classic framework based on VQ-VAE~\cite{van+:2017neural} and GPT to synthesize human motion from textual descriptions, which achieved comparable performances than diffusion-based approaches. MotionGPT~\cite{jiang+:2023motiongpt} is a uniform motion-language generative pre-trained model, which treats human motion as a tokenized foreign language, enabling seamless integration of natural language models into motion-relevant generation for diverse motion tasks.
\section{Conclusions and Future Work}
\label{sec:conclusion}
In this work, we introduced a novel approach to enhance continuous sign language translation from text to sign pose sequences by explicitly modeling geometrical constraints and anatomical consistency. Specifically, parent-relative reweighting, which dynamically emphasizes critical joints during training, significantly improves the naturalness and flexibility of generated finger and hand articulations. Additionally, we integrated bone-length and bone-pose losses to maintain accurate and anatomically consistent joint positions and motions throughout generated sequences. \looseness=-1

Experimental evaluations on the PHOENIX-2014 benchmark dataset demonstrate substantial improvements over previous methods. Our approach reduced the performance discrepancy to the theoretical upper bound (back-translation on ground truth) by 56.51\%, highlighting significant enhancements in translation quality. Quantitative analyses further indicated notable improvements in anatomical consistency, achieving an 18.76\% reduction in bone length discrepancies and a 5.48\% reduction in movement variance differences compared to the baseline. Qualitatively, our method consistently generated more realistic and expressive sign sequences with smoother and anatomically plausible transitions. \looseness=-1

Future work will explore integrating additional sign-related features, such as facial expressions and non-manual markers, to further enhance the expressiveness and realism of sign language generation. Additionally, incorporating generative adversarial networks (GANs) or other discriminative training techniques could help improve the authenticity and fluidity of generated sequences. Extending our approach to diverse sign languages and datasets will also be essential for validating its generalizability and real-world applicability. \looseness=-1
{
    \small
    \bibliographystyle{ieeenat_fullname}
    \bibliography{main}
}

\end{document}